\documentclass{article}
\usepackage[utf8]{inputenc}
\usepackage{graphicx}
\usepackage{amsmath, amssymb}
\usepackage{algorithm}
\usepackage{algorithmic}
\usepackage{hyperref}
\usepackage{bm}
\usepackage{booktabs}
\usepackage{float}
\usepackage{caption}
\usepackage{subcaption}
\usepackage{geometry}
\usepackage{soul,xcolor}
\usepackage{tikz}
\usepackage{bbm}
\usepackage{multirow}
\usepackage{titlesec}

\usepackage{multicol,lipsum}
\usetikzlibrary{shapes.geometric, arrows, positioning, fit, calc}
\title{Superposition in Transformers: A Novel Way of Building Mixture of Experts}
\author{
    Ayoub Ben Chaliah \\
    \small \href{mailto:ayoub1benchaliah@gmail.com}{ayoub1benchaliah@gmail.com}
    \and
    Hela Dellagi \\
    \small \href{mailto:hela.dellagi@outlook.com}{hela.dellagi@outlook.com}
}
\date{\today}

\begin{document}
\setstcolor{red}
\maketitle
\begin{multicols}{2}
\begin{abstract}
Catastrophic forgetting remains a major challenge when adapting large language models (LLMs) to new tasks or domains. Conventional fine-tuning often overwrites existing knowledge, causing performance degradation on original tasks. We introduce \textit{Superposition in Transformers}, a novel architecture that leverages autoencoders to superimpose the hidden representations of a base model and a fine-tuned model within a shared parameter space. By using B-spline-based blending coefficients and autoencoders that adaptively reconstruct hidden states based on the input data distribution, our method effectively mitigates catastrophic forgetting and enables a new paradigm of “in-model” superposition. This approach preserves original model capabilities while allowing compact domain-specific expertise to be added, and it supports dynamic switching between model states during inference.
\end{abstract}

\section{Introduction}
\label{submission}
Large language models (LLMs) such as GPT-3~\cite{brown2020language} and GPT-4~\cite{openai2023gpt4} have shown remarkable performance on various language tasks. However, adapting these models to new tasks or domains often leads to \textbf{catastrophic forgetting}, where newly learned information overwrites older knowledge~\cite{mccloskey1989catastrophic}. This challenge becomes critical in continual learning or domain adaptation settings, where preserving performance on original tasks is essential~\cite{rusu2016progressive, kirkpatrick2017overcoming}.\\
Most existing Mixture of Experts (MoE)~\cite{jacobs1991adaptive}  methods expand parameter count significantly by introducing distinct expert modules and gating networks~\cite{shazeer2017outrageously,Yun2024moe}. Parameter-efficient fine-tuning approaches such as Adapters~\cite{houlsby2019parameter} or LoRA~\cite{hu2021lora} often rely on modifying or appending new weights to the base model~\cite{wang2022adamix, Zadouri2023pushing, wu2023omni}. In contrast, our method produces a \emph{single} merged set of parameters—via B-spline blending—while training autoencoders to reconstruct the original hidden states on demand. During training, both the base and fine-tuned model parameters are accessed in a frozen manner to generate hidden states, but in the end, only the new blended model parameters (plus the autoencoders) remain. This design merges the capabilities of two experts into one compact parameter space and mitigates catastrophic forgetting by preserving each model's knowledge in a superposed, reconstructable form.

The contributions of this paper are as follows:\\
 \textbf{Autoencoder-Based Superposition :} We demonstrate how autoencoders can reconstruct hidden states from either the base or fine-tuned model, effectively gating the representation space to prevent catastrophic forgetting.\\
 \textbf{Jointly Learned B-Spline Blending :} By training blending coefficients jointly with the autoencoders, we find a balance that preserves each model’s capabilities while introducing minimal new parameters.\\
\textbf{Parameter Efficiency and Practicality:} This approach retains the majority of each model’s parameters in frozen form and only learns relatively small auxiliary modules, making it feasible for real-world use.\\
\textbf{Future Directions:} We discuss possible expansions of this framework, such as merging specialized experts for chain-of-thought reasoning or leveraging token-level signals to dynamically switch hidden states within the same forward pass.

\section{Background and Related Work}

Our work lies at the intersection of several key research areas in neural networks, including Mixture of Experts (MoE) models, parameter-efficient transfer learing, and neural network compression. Additionally, we draw from studies on superposition and polysemantic neurons in language models.

\textbf{Mixture of Experts. }
MoE models dynamically allocates different experts to different inputs, improving capacity and specialization. While effective, MoE methods often increase parameter count and rely on gating networks~\cite{Yun2024moe}. Our method similarly aims to leverage multiple “expert” models but does so via a shared parameter space and lightweight autoencoder modules.

\textbf{Parameter-Efficient Transfer Learning .}
Recent techniques like Adapters~\cite{houlsby2019parameter} and LoRA~\cite{hu2021lora} reduce the cost of fine-tuning large models by injecting small trainable components. We differ in that we freeze both the base and fine-tuned models; instead of adding new layers directly on top of them, we \emph{blend their hidden states} and \emph{reconstruct} them adaptively. This approach can be viewed as orthogonal to Adapters or LoRA, potentially combining well with these methods.

\textbf{Superposition and Polysemantic Neurons .}
In neural networks, \textit{superposition} refers to efficiently encoding multiple features or tasks within the same parameter space~\cite{Elhage2022toy}. Large LLMs often exhibit \textbf{polysemantic neurons}—units that respond to multiple unrelated concepts~\cite{olah2020zoom}. Recent research into feature interpretability has shown that sparse autoencoders (SAEs) can transform model activations into more interpretable feature spaces, helping to disentangle polysemantic neurons into monosemantic units ~\cite{ Bricken2023toward,OMahony2023disentangling,Huben2024sparse}. By leveraging these learned features, researchers have identified significant similarities across latent spaces in different LLMs, suggesting a \textbf{shared representational structure} ~\cite{Bricken2023toward}.
Our approach promotes polysemanticity by forcing the network to encode the states of two distinct models in overlapping neurons, constrained by autoencoder bottlenecks.

\textbf{Neural Network Compression and Model Merging .}
Prior methods for merging or compressing models~\cite{li2020train,matena2022merging} generally aim to reduce memory overhead. By contrast, our framework focuses on \textit{preserving both original models} in a single parameter space without overwriting. The added overhead is small (primarily the autoencoders and the blending coefficients) compared to training or storing two separate models.

\textbf{Preventing Catastrophic Forgetting .}
Preventing catastrophic forgetting, where models lose previously learned knowledge when fine-tuned on new tasks, has been a critical area of research in transfer learning \cite{kirkpatrick2017overcoming}. Our method addresses this by freezing the base and fine-tuned models, thus preserving their original knowledge. The B-spline blending coefficients and autoencoders provide a flexible mechanism for combining knowledge from both models without overwriting their respective features. This approach shares similarities with knowledge distillation techniques but focuses on preserving knowledge across layers rather than compressing it into a smaller model.

\section{Proposed Method}

\subsection{Overview}
We merge a base model \(M_{\text{base}}\) and a fine-tuned model \(M_{\text{fine}}\) into a single “superposed” model \(M_{\text{merged}}\). To do this, we:\\

\textbf{Blend Hidden States per Layer} using \(\alpha(l)\), which we compute via B-spline interpolation.\\
\textbf{Insert Autoencoders} at selected layers to reconstruct the blended states either as \(\mathbf{h}^{\text{base}}\) or \(\mathbf{h}^{\text{fine}}\), based on input labels or domain cues.\\
\textbf{Train Jointly} the B-spline control points, layer biases, and autoencoder parameters so that each layer’s \(\alpha(l)\) best preserves domain-specific details when guided by the autoencoder reconstruction loss.

\subsection{Blending Model Weights Using B-Splines}

\subsubsection{Motivation}
Averaging or naively mixing model weights can degrade performance, as it does not adapt to which features are critical to each model. Instead, we \emph{blend hidden states} directly, allowing each layer to remain mostly intact (frozen). The \(\alpha\)-values are learned with a mechanism that encourages smooth transitions across layers.

\subsubsection{Formulation}

For layer \(l\), let \(\mathbf{h}_l^{\text{base}}\) and \(\mathbf{h}_l^{\text{fine}}\) be the hidden states from the base and fine-tuned models, respectively. We define
\[
\mathbf{h}_l = (1 - \alpha(l))\,\mathbf{h}_l^{\text{base}} + \alpha(l)\,\mathbf{h}_l^{\text{fine}}, 
\]
\[
\alpha(l) = \mathrm{clamp}\!\Bigl(\sum_{i=1}^{N} c_i\,B_{i,k}(l) + b_l,\;0,\;1\Bigr)
\]

where \(\{c_i\}\) are trainable control points for a B-spline of degree \(k\) and  \(b_l\) is a layer-specific bias term. \\
The B-spline basis functions \(B_{i,k}(\cdot)\) ensure smoothly varying \(\alpha(l)\).
We freeze the weights of both \(M_{\text{base}}\) and \(M_{\text{fine}}\), focusing on learning only \(\{c_i\}\), \(\{b_l\}\), and the autoencoders.

\subsection{Merged Model Architecture}

\subsubsection{Merging Models Post-Training}
After training, one option is to produce a single model whose parameters reflect a final \textit{hard merge} of the two original models:
\[
\theta_l = (1 - \alpha(l))\,\theta_l^{\text{base}} + \alpha(l)\,\theta_l^{\text{fine}},
\]
where \(\theta_l^{\text{base}}\) and \(\theta_l^{\text{fine}}\) are the layer parameters. This yields a standalone model with minimal overhead. Alternatively, one may keep the B-spline plus autoencoders as a gating system that dynamically adapts to inputs.

\subsubsection{Forward Pass with Merged Model}
During inference, the model computes hidden states for each layer via blended embeddings and self-attention (incorporating \(\alpha(l)\)) and optionally applies an autoencoder to refine \(\mathbf{h}_l\), reconstructing the hidden states that best suit the input’s domain.
This structure preserves each model’s domain knowledge while introducing minimal overhead.

\subsection{Autoencoders for State Reconstruction}

\subsubsection{Architecture}
At each selected layer \(l\), an autoencoder compresses the blended hidden state \(\mathbf{h}_l\) into a bottleneck and reconstructs it: $\mathbf{z}_l = \mathrm{Encoder}(\mathbf{h}_l)$ and $\hat{\mathbf{h}}_l = \mathrm{Decoder}(\mathbf{z}_l)$.

We train it to match \(\hat{\mathbf{h}}_l\) with either \(\mathbf{h}_l^{\text{base}}\) or \(\mathbf{h}_l^{\text{fine}}\). We \textit{skip} embedding layers and the final layer norm to maintain alignment with the original model heads.

\subsubsection{Minimizing Information Loss}
Because \(\alpha\) and the autoencoder parameters are trained jointly, \(\alpha\) naturally adjusts so that the autoencoder can accurately reconstruct either \(\mathbf{h}_l^{\text{base}}\) or \(\mathbf{h}_l^{\text{fine}}\). This synergy ensures that if reconstruction error is high, \(\alpha(l)\) shifts to a region that better preserves salient features from the respective domain.

\subsubsection{Role of Autoencoders in Encouraging Polysemanticity}
Each autoencoder’s bottleneck forces the network to encode crucial details from both models within a limited dimension. This encourages \textbf{polysemantic} neurons, since the same hidden units may now carry signals relevant to both tasks. A narrower bottleneck heightens polysemantic pressure but can also degrade domain-specific detail; a wider bottleneck reduces polysemantic pressure but can maintain more domain specificity.

\subsubsection{Extending to a 2D-alpha Model (Optional)}
One can extend \(\alpha(l)\) to a vector for per-dimension blending. In this scenario, local features (extracted by convolution layers) and global features (captured by a low-rank adapter) are combined in the autoencoder. The 2D-alpha approach delves deeper into exploring the impact of the autoencoder on the polysemantic nature of neurons within a transformer block. While more expressive, it also increases complexity and may demand careful tuning to avoid overfitting.

\begin{figure*}[t]
    \centering
\resizebox{0.6\textwidth}{!}{
\begin{tikzpicture}[
    block/.style={rectangle, draw, minimum width=2.5cm, minimum height=1cm, align=center, rounded corners},
    smallblock/.style={rectangle, draw, minimum width=1.8cm, minimum height=0.7cm, align=center, rounded corners, font=\footnotesize},
    arrow/.style={->, thick},
    sum/.style={circle, draw, inner sep=0.05cm}
]
\colorlet{decodercolor}{orange!20}
\colorlet{subdecodercolor}{green!20}
\colorlet{reconstructioncolor}{blue!20}

\node[block, fill=decodercolor] (decoder1) at (0,0) {Blending hidden states};
\node[block, fill=reconstructioncolor, below=0.2cm of decoder1] (recon1) {Autoencoder};
\node[below=0.1cm of recon1, font=\small] (trans1) {Transformer 1};
\node[fit=(decoder1) (recon1) (trans1), inner sep=3pt, draw] (block1) {};
\node[below=0.1cm of block1,xshift=-0.3cm, font=\small] (out1) {$\hat{h}_1$};

\node[below=0.5cm of block1] (dots) {\vdots};

\node[below=-0.2cm of dots,xshift=-0.3cm, font=\small] (outl11) {$\hat{h}_{l-1}$};

\node[block, fill=decodercolor, below=0.5cm of dots] (decoderl) {Blending hidden states};
\node[block, fill=reconstructioncolor, below=0.2cm of decoderl] (reconl) {Autoencoder};
\node[below=0.1cm of reconl, font=\small] (transl) {Transformer $l$};
\node[fit=(decoderl) (reconl) (transl), inner sep=3pt, draw] (zoom) {};
\node[below=0.05cm of zoom,xshift=-0.3cm, font=\small] (outl) {$\hat{h}_l$};

\node[below=0.5cm of zoom] (dots2) {\vdots};

\node[block, fill=decodercolor, below=0.5cm of dots2] (decoderm) {Blending hidden states};
\node[block, fill=reconstructioncolor, below=0.2cm of decoderm] (reconm) {Autoencoder};
\node[below=0.1cm of reconm, font=\small] (transm) {Transformer 12};
\node[fit=(decoderm) (reconm) (transm), inner sep=3pt, draw] (block12) {};
\node[below=0.1cm of block12,xshift=-0.3cm, font=\small] (outm) {$\hat{h}_{12}$};

\draw[arrow] (decoder1) -- (recon1);
\draw[arrow] (decoderl) -- (reconl);
\draw[arrow] (decoderm) -- (reconm);

\draw[arrow] (block1) -- (dots);
\draw[arrow] (dots) -- (zoom);
\draw[arrow] (zoom) -- (dots2);
\draw[arrow] (dots2) -- (block12);

\begin{scope}[xshift=7.9cm, yshift=-0.9cm, local bounding box=zoomed_decoder]
    \node[above=0.5cm, font=\bfseries] (title_decoder) {Blending Hidden States};
    \node[smallblock, below=0.3cm of title_decoder] (hl1) {$h_{l-1}$};
    \node[smallblock,fill=subdecodercolor, below left=0.5cm and -0.5cm of hl1] (hlbase) {Base Hidden State\\$h_l^{\text{base}}$};
    \node[smallblock,fill=subdecodercolor, below right=0.5cm and -0.5cm of hl1] (hlfine) {Fine-tuned Hidden State\\$h_l^{\text{fine}}$};
    \node[smallblock, below=2cm of hl1] (blended) {Blended Hidden State $h_l = (1 - \alpha(l)) h_l^{\text{base}} + \alpha(l) h_l^{\text{fine}}$\\ $\alpha(l)$ can be a scalar or vector depending on the architecture} ;
    
    \draw[arrow] (hl1) -- (hlbase);
    \draw[arrow] (hl1) -- (hlfine);
    \draw[arrow] (hlbase) -- (blended);
    \draw[arrow] (hlfine) -- (blended);
    
    \node[fit=(title_decoder) (hl1) (hlbase) (hlfine) (blended) , inner sep=20pt, draw=black, dashed] (decoder_box) {};
\end{scope}

\begin{scope}[xshift=8cm, yshift=-7.5cm, local bounding box=zoomed_reconstruction]
    \node[above left=0.4cm, font=\bfseries] (title) {};
    \node[font=\bfseries, align=center] at (0.8, 0.5) (movabletext) {Autoencoder Block};

    \node[smallblock, below=0.2cm of title ] (el) {Encoder};
    \node[smallblock, right=0.2cm of el] (lr) {Low Rank Adapter\\(Optional)};
    \node[smallblock, below =1cm of el] (zl) {Decoder};
    \node[circle, draw, below=0.6cm of el, minimum size=0.2cm, inner sep=0pt] (plus) {\footnotesize +};
    \node[below=0.5cm of zl, align=center] (eq) {Reconstructed State};
    
    \draw[arrow] (zl) -- (eq);
    \draw[-] (el) -- (plus); 
    \draw[-] (plus) -- (zl); 
    \draw[-] (lr.south) |- (plus);
    \node[fit=(title) (zl) (eq) (lr), inner sep=20pt, draw=black, dashed] (recon_box) {};
\end{scope}

\draw[dashed] (decoderl.east) -- ++(1,0) |- (zoomed_decoder);
\draw[dashed] (reconl.east) -- ++(1,0) |- (zoomed_reconstruction);

\end{tikzpicture}}

    \caption{Overview of a GPT-2 Merged Model Architecture. }
    \label{fig:architecture}
\end{figure*}
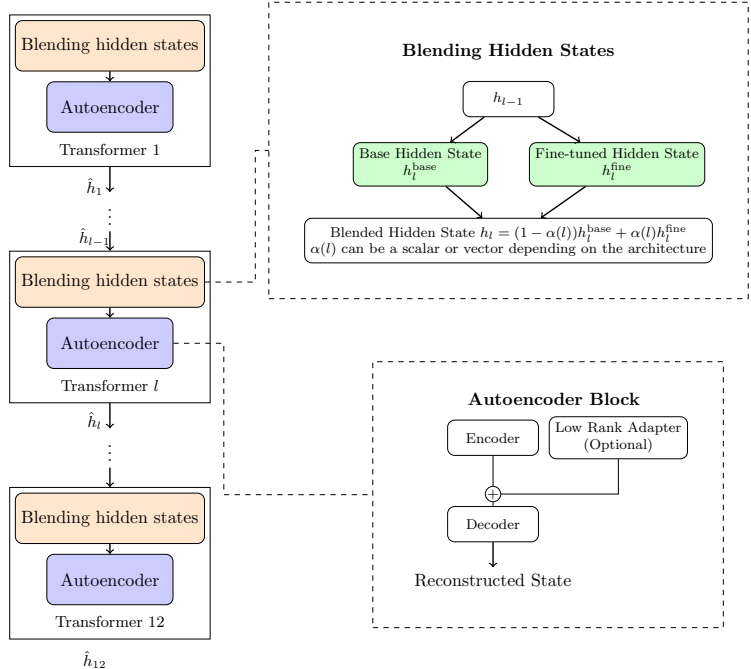

\subsection{Training Procedure}

\subsubsection{Objectives}

We optimize two core objectives: a standard language modeling loss $\mathcal{L}_{\mathrm{LM}}$, preserving fluency and coherence and the reconstruction error between $\hat{\mathbf{h}}_l$ and the target hidden state $\mathbf{h}_l^{\text{base}}$ or $\mathbf{h}_l^{\text{fine}}$. An additional alpha regularization term \(\mathcal{L}_{\mathrm{AlphaReg}}\) may be applied to keep \(\alpha\) smooth or centered.

\subsubsection{Loss Functions}
We generally use:
$$
\mathcal{L} \;=\; \lambda_{\mathrm{LM}}\,\mathcal{L}_{\mathrm{LM}}
\;+\;\lambda_{\mathrm{Recon}}\,\mathcal{L}_{\mathrm{Recon}}
\;+\;\lambda_{\mathrm{Alpha}}\,\mathcal{L}_{\mathrm{AlphaReg}}$$
where \(\lambda_{\mathrm{LM}}\), \(\lambda_{\mathrm{Recon}}\), and \(\lambda_{\mathrm{Alpha}}\) balance the relative importance of each component. \(\mathcal{L}_{\mathrm{Recon}}\) is typically Mean Squared Error (MSE) or L2 distance over the hidden states. $\mathcal{L}_{\mathrm{AlphaReg}}$ encourages the control points of the
B-spline-based blending coefficients to adhere to desirable properties. It ensures smooth and interpretable
blending, avoiding overfitting to noisy representations. 
\subsubsection{Optimization}
We freeze all parameters of \(M_{\text{base}}\) and \(M_{\text{fine}}\). The trainable variables are the B-spline control points \(\{c_i\}\) and biases \(\{b_l\}\) as well as the autoencoder weights and biases per selected layer.\\
By iterating through mini-batches, we compute hidden states from both models (frozen), blend them via \(\alpha(l)\) to produce \(\mathbf{h}_l\) and use an autoencoder to reconstruct \(\mathbf{h}_l\) into either \(\mathbf{h}_l^{\text{base}}\) or \(\mathbf{h}_l^{\text{fine}}\). We then backpropagate the reconstruction error to update \(\alpha\) and the autoencoder parameters.
The final forward pass for language modeling also contributes to \(\mathcal{L}_{\mathrm{LM}}\), ensuring that fluency is maintained.

\subsubsection{Role of Labels}
Labels indicating whether an input belongs to the base or fine-tuned domain can guide which hidden state the autoencoder targets. While optional in principle, these labels can significantly speed up convergence and improve accuracy of reconstruction for each domain.

\section{Experiments and Results}
\label{sec:experiments}
In this section, we evaluate the proposed merged model, by integrating a base GPT-2~\cite{Radford2019language} model and a fine-tuned GPT-2 model trained on French text. We focus on demonstrating how the autoencoders enable the superposition of transformer blocks from different fine-tunes, allowing the merged model to effectively combine representations from both models. Our analysis includes performance metrics, hidden state reconstruction and the emergence of polysemantic neurons.

\subsection{Experimental Setup}

We compare a Base Model (GPT-2 trained on English data), a Fine-Tuned Model (GPT-2 fine-tuned on a French corpus) and the Merged Model which combines the base and fine-tuned models using layer-wise blending with learned $\alpha$ values and incorporates autoencoders to enable superposition.
For evaluation, we used two datasets, an english dataset that contains a subset of the GPT-2 training data (6,000 samples repeated with range(3) in the 1D experiment and repeated with range(2) in the 2D experiment) and a french dataset (18,000 French Wikipedia articles for the 1D model and 12,000 French Wikipedia articles for the 2D model). 10\% of the combined samples are reserved for validation.\\
In the case of the 1D-alpha GPT2 model the number of parameters per autoencoder is: $4 \times \text{dimension of the hidden state}\times \text{bottleneck size}  + \text{dimension of the hidden state}+ \text{bottleneck size} $. The bottleneck size used in the 1D-alpha model experiment is 576, resulting in approximately 1.7 million additional parameters per autoencoder.

\subsection{Results of the 1D model}

\subsubsection{Perplexity per language}

Tables 1 compares the perplexity of the models using French and English inputs :
The perplexity results show distinct performance characteristics for the models. For French inputs, the fine-tuned model has a significantly lower perplexity (29.57) compared to the base model (132.02), indicating its suitability for French. Conversely, the base model performs best on English inputs with a lower perplexity (28.88), suggesting it was primarily optimized for English.

\noindent The merged model demonstrates closer perplexity to the base model in English (43.14 vs. 28.88) and is comparable to the fine-tuned model in French (33.20 vs. 29.57), indicating its effectiveness in bridging the performance across both languages.

\begin{table}[H]
\centering
\resizebox{\columnwidth}{!}{%
\begin{tabular}{lcc}
\toprule
\textbf{Model}               & \textbf{Language} & \textbf{Perplexity} \\
\midrule
Base Model                   & English           & 28.88 \\ 
                             & French            & 132.02 \\
\midrule
Fine-Tuned Model             & English           & 54.42 \\ 
                             & French            & 29.57 \\ 
\midrule
Merged Model                 & English           & 43.14 \\ 
                             & French            & 33.20 \\
\bottomrule
\end{tabular}%
}
\caption{Perplexity of the models on the evaluation datasets.}
\end{table}
\noindent The merged model demonstrates closer perplexity to the base model in English (43.14 vs. 28.88) and is comparable to the fine-tuned model in French (33.20 vs. 29.57), indicating its effectiveness in bridging the performance across both languages.

\subsubsection{Overall Performance}

To empirically demonstrate the advantages of our \textit{Superposition in Transformer} technique, we conducted experiments comparing Superposition Merging with linear interpolation and task arithmetic techniques. The perplexity of the autoencoder-merged model (\textbf{M-PPL=47.01}) was markedly lower than that of the linearly interpolated model (\textbf{I-PPL=60.29}) and the task arithmetic model (\textbf{I-PPL=61.30}) during the last epoch of training, suggesting a higher confidence in predicting the next token. The merged model initially showed higher perplexity than both task arithmetic and linear interpolation models but then achieved lower values as training progresses (figure 2). \\
Similarly, the next-token prediction accuracy was also improved with Superposition Merging \\(\textbf{M-Acc=0.3270}) compared to linear interpolation (\textbf{I-Acc=0.3039}) and task arithmetic (\textbf{I-Acc=0.2957}).\\

\begin{figure}[H]
\centering
\includegraphics[width=0.45\textwidth]{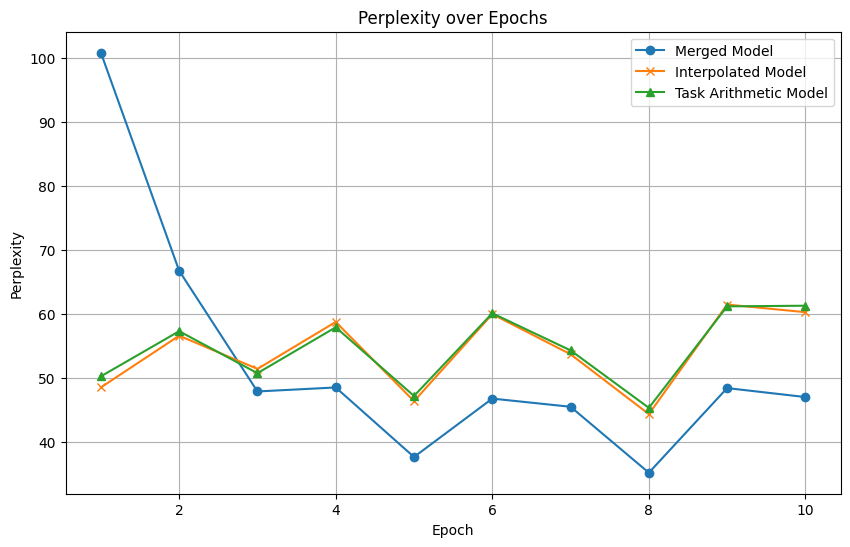}
\caption{Perplexity evolution across epochs for different merging methods.}
\end{figure}

Additionnaly, we examined the Jensen-Shannon Divergence JSD values over different epochs during training. The JSD drops sharply from \textbf{M-JSD=82.9} to \textbf{M-JSD=45.5} in the first two epochs. This drastic reduction indicates that the merged model's output distribution quickly diverges from the initial average of the base and fine-tuned models. After this initial drop, the JSD values fluctuated between approximately \textbf{M-JSD=36.5} and \textbf{M-JSD=44.2}, suggesting that the merged model settles into a relatively stable output distribution that is distinct from the simple average.

These quantitative results strongly suggest that Superposition Merging, facilitated by the learning capabilities of autoencoders, offers a more robust and effective method for combining the strengths of independently trained models. Unlike traditional methods that can lead to a homogenization of expertise, our approach shows promising potential for achieving a genuine superposition, where the merged model can effectively leverage the unique skills of each expert model within their respective domains. This opens up exciting possibilities for creating more versatile and powerful AI systems by intelligently combining specialized knowledge.

\begin{figure*}[t] \centering \includegraphics[width=0.9\textwidth]{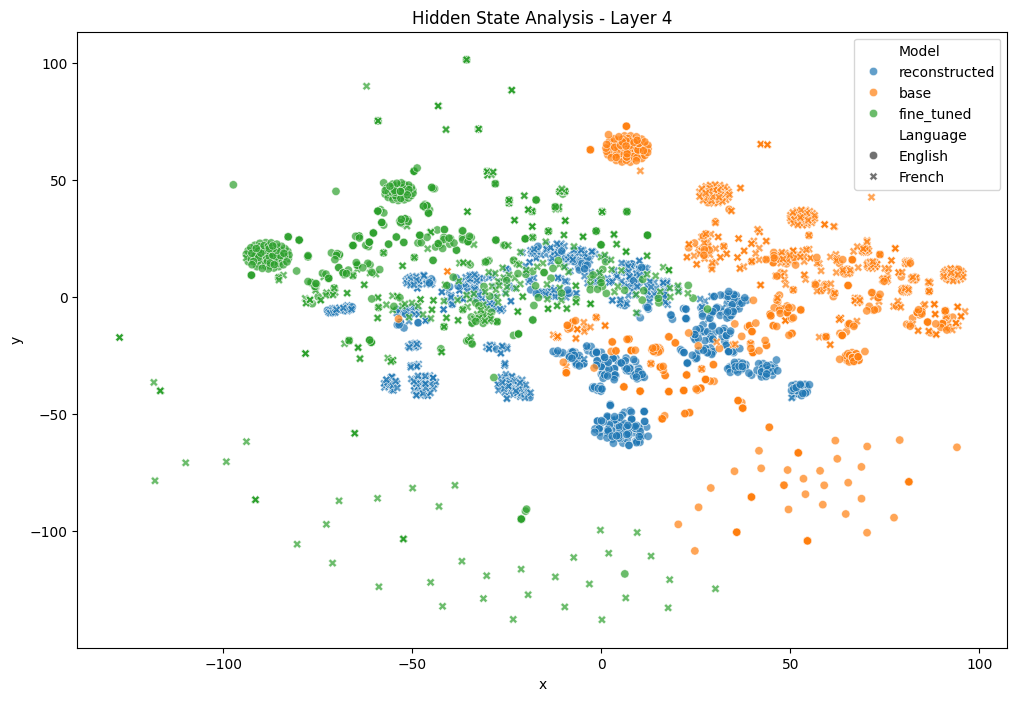} \caption{t-SNE visualization of layer 4 hidden states from the merged model and expert models for English and French inputs (2D-alpha).} \end{figure*}

\subsection{Results of the 2D model}
\subsubsection{Hidden States in Layer 4}

To evaluate how the autoencoders enable the adaptive blending of representations in the 2D-alpha model, a t-SNE analysis is conducted on the hidden states from layer 4. Similar to the 1D-alpha model analysis, figure 3 effectively illustrates the model's capability to align reconstructed hidden states with the appropriate model based on input language.
Specifically, for English inputs, the reconstructed states from the merged model are closely aligned with the base model, indicated by the use of "$\circ$" markers in the graph. Conversely, for French inputs, these states align more closely with the fine-tuned model, which is depicted using "x" markers. This distinction visually underscores the model’s adaptive capacity to toggle between the base and fine-tuned model states depending on the language context

\subsubsection{Neuron Utilization and Polysemantic Neurons}

\begin{table}[H]
\centering
\resizebox{\columnwidth}{!}{%
\begin{tabular}{lcc}
\toprule
\textbf{Model} & \textbf{Sparsity (\%)} & \textbf{Polysemantic Neurons (\%)} \\
\midrule
Base Model       & 0.0166 & 1.8229 \\
Fine-Tuned Model & 0.0154 & 2.0833 \\
Merged Model     & 0.0179 & 3.3854 \\
\bottomrule
\end{tabular}%
}
\caption{Sparsity and proportion of polysemantic neurons in layer 4 of the models.}
\end{table}

The 2D-alpha model demonstrates enhanced neuron utilization (table 2). The increase in sparsity is minimal and potentially arbitrary due to the model's small size.\\
The percentage of polysemantic neurons is computed using a threshold value of 0.05. This threshold delineates neurons based on the normalized difference in their average activations across different language contexts (e.g., English vs. French). Neurons with a normalized difference below this threshold are considered polysemantic because they react similarly across linguistic contexts, suggesting a shared semantic load. However, using such a fixed threshold might not reliably capture the full spectrum of neuron functionality or their sensitivity to different contexts.
Given the potential limitations of using a fixed threshold for identifying polysemantic neurons, the comparative analysis of neuron diversity and activation employs a more dynamic approach. 
\subsubsection{Comparative Analysis of Neuron Diversity and Activation}
To gain deeper insights into how the 2D-alpha merged model utilizes neurons compared to the base and fine-tuned models, we conducted an analysis of neuron diversity and activation across layers. Instead of defining polysemicity with a single threshold, this analysis evaluates the diversity of neuron responses through a MiniBatchKMeans clustering algorithm,. The diversity is quantified using entropy measures derived from the distribution of neurons across clusters formed based on their activation patterns. This method provides a deeper insight into how neurons respond to different linguistic inputs, reflecting a broader range of neuron functionality than can be detected by a simple threshold-based method.\\ \\
\textbf{Note:}  In the 2D-alpha model, the autoencoders were used starting from layer 4 until layer 10. Figures 4 and 5 clearly illustrate the contrast between layer 3 and the subsequent layers 4, 5, and 6.\\ \\
 The reduction in diversity metrics (figure 4) shows that neurons encode overlapping or generalized features. While layers without (e.g. layer 3) autoencoders retain higher diversity due to the presence of specialized neurons, the autoencoder forces neurons to unify and encode features relevant across multiple tasks or contexts. This suggests a shift from task-specificity to multi-purpose functionality—a hallmark of polysemantic neurons.\\
 
\begin{figure}[H] \centering \includegraphics[width=0.45\textwidth]{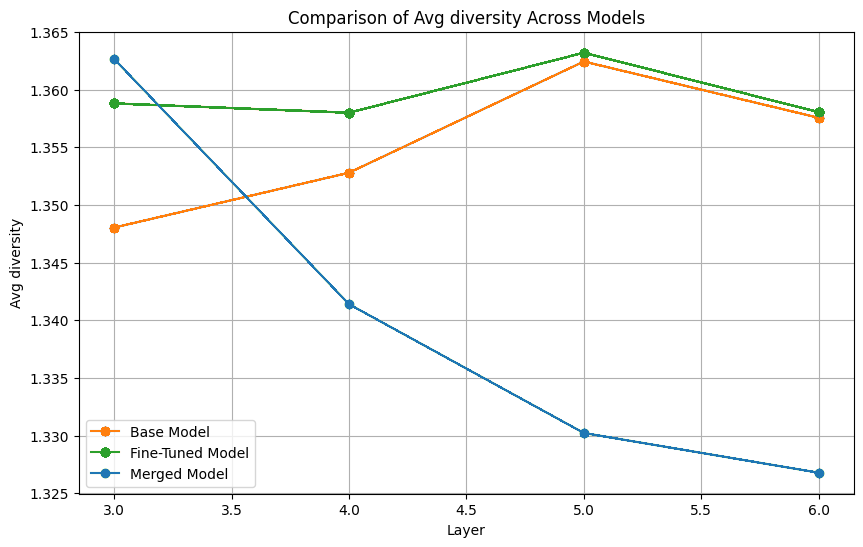} \caption{Comparison of average neuron diversity across layers for the base, fine-tuned, and merged models.} \end{figure}

\begin{figure}[H] \centering \includegraphics[width=0.45\textwidth]{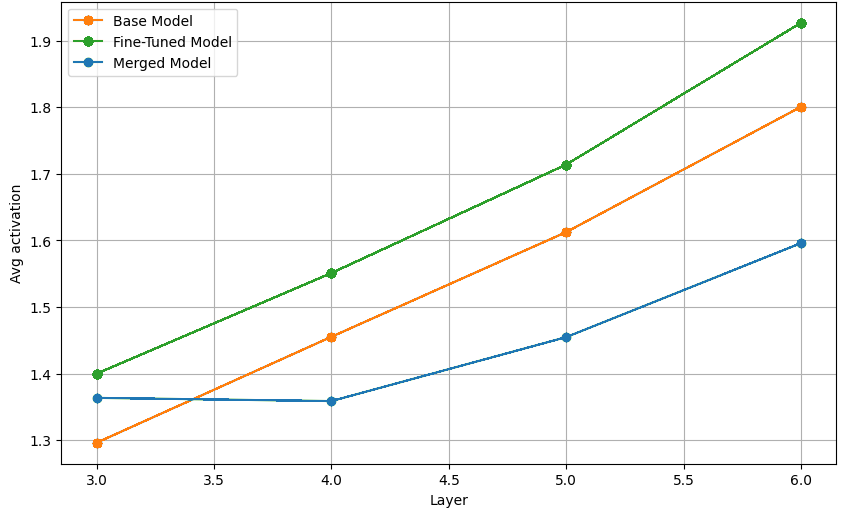} \caption{Comparison of mean neuron activation across layers for the base, fine-tuned, and merged models.} \end{figure}

\noindent The lower average activation magnitudes observed in pre-autoencoder layers reflect a shift towards encoding distributed, low-magnitude representations. This subtle encoding strategy ensures that the autoencoder can reconstruct activations effectively while maintaining compactness. Distributed, low-magnitude activations are more likely to encode polysemantic features, as they provide flexibility and reuse across tasks.

\section{Limitations}

While our method demonstrates promising results, several limitations should be noted. Currently, this approach might does not support dynamic state switching within a single input. Achieving this may require additional mechanisms, such as token-level state signaling, which remains unexplored. Additionally, our experiments focus solely on merging two models—one base model and one fine-tuned model. Extending this approach to handle more than two models, such as combining multiple experts or integrating diverse knowledge domains, may pose scalability challenges and require modifications to the blending and reconstruction processes.

\section{Future Work}

 \textbf{Tuning Bottleneck Hyperparameters}: Experimenting with different bottleneck sizes can help balance the trade-off between generalization and task-specificity. \\
 \textbf{Analyzing Cross-Layer Interactions}: Investigate how the presence of an autoencoder in one layer influences subsequent layers. Does the polysemanticity induced in one block propagate through the network?\\
 \textbf{Enable Language Switching Within a Single Context}: A promising direction is enabling the model to alternate between languages within a single context, such as generating output that seamlessly switches between English and French while maintaining coherence and fluency. This could be particularly valuable for tasks requiring multilingual communication, such as translation or conversational agents operating in diverse linguistic environments. Special tokens or contextual cues could signal the model to adjust its representations dynamically, ensuring that the transitions between languages preserve the original intent and meaning of the sentence.\\
 \textbf{Merging Specialized Experts}: We plan to explore merging a symbolic reasoning expert with a domain knowledge expert. This would enable the model to switch between symbolic processing and contextual understanding dynamically. This idea is analogous to the language switching above, as it allows seamless transitions between different modes of cognition while preserving the overarching context.\\
\textbf{Single Pass Generation}: Similar to the language switching experiment, special tokens or cues could be used to signal the model to switch states within a single generation sequence.\\ \\
 \textbf{Example Scenario}:
    \begin{itemize}
        \item \textit{Input}: "Solve for $x$: $2x + 3 = 7$. Then, explain the significance of the solution in real-world applications."
        \item \textit{Expected Behavior}:
        \begin{itemize}
            \item \textbf{First Part (Symbolic Reasoning)}: The model uses the symbolic reasoning expert to solve the equation.
            \item \textbf{Second Part (Domain Explanation)}: The model switches to the domain knowledge expert to provide an explanation.
        \end{itemize}
    \end{itemize}

\section{Conclusion}
The proposed method represents a significant advancement in enabling large language models (LLMs) to integrate knowledge from multiple domains or tasks without sacrificing foundational capabilities or requiring extensive parameter growth. By leveraging autoencoders and blending mechanisms, the approach facilitates efficient knowledge integration making it a practical and resource-efficient enhancement to LLMs.\\
Experimental results highlight improvements in perplexity reduction, reconstruction accuracy, and hidden state alignment with base and fine-tuned models. Techniques like t-SNE visualizations further validate the model's ability to adapt representations dynamically. This adaptability enables seamless processing of diverse inputs and enhances multi-domain integration.\\
Beyond merging two models, this method promises broader applications, such as multilingual processing, seamless language switching, and integrating symbolic reasoning with domain knowledge.\\
Looking ahead, this approach paves the way for the development of models capable of integrating diverse skills and knowledge domains, representing a step toward more flexible, powerful, and general-purpose AI systems. The potential applications are vast and exciting.

\end{multicols}


\appendix
\onecolumn

\section*{\centering\Huge\bfseries Appendix}
\addcontentsline{toc}{section}{Appendix} 
\setcounter{section}{0} 

\section{Training Algorithms}

\subsection{Loss Functions}

The losses involved in the different architectures combines:

\paragraph{Reconstruction Loss $\mathcal{L}_{\text{Recon}}$} 
This loss measures how well the autoencoders reconstruct the blended hidden states. It uses both Mean Squared Error (MSE) and the L2 distance between the reconstructed hidden states and the target hidden states. 
\begin{equation}
\mathcal{L}_{\text{Recon}} = \sum_{l=1}^L\left[ \lambda_{\text{MSE}}\mathbb{E}\left[ \left\| \hat{\mathbf{h}}_l - \mathbf{h}_l^{\text{target}} \right\|_2^2\right] + \lambda_{\text{L2}} \left\| \hat{\mathbf{h}}_l - \mathbf{h}_l^{\text{target}} \right\|_2\right] 
\end{equation}
L represents the number of layers in the transformer model.
\paragraph{Language Modeling Loss $\mathcal{L}_{\text{LM}}$} This loss measures the standard language modeling objective using cross-entropy between the predicted logits and the next token in the sequence.
\begin{equation}
\mathcal{L}_{\text{LM}} = \mathbb{E}\left[ -\sum_{t} \log P_{\theta}(w_t | w_{<t}) \right]
\end{equation}

\paragraph{Alpha Regularization Loss $\mathcal{L}_{\text{AlphaReg}}$}
The alpha regularization loss encourages the control points of the B-spline-based blending coefficients to adhere to desirable properties. It ensures smooth and interpretable blending, avoiding overfitting to noisy representations. It includes three components:

\begin{itemize}
    \item \textbf{Smoothness Loss}: Ensures that adjacent control points are close to each other, minimizing abrupt changes in blending behavior. \begin{equation}\mathcal{L}_{\text{Smoothness}} = \sum_{i=1}^{N-1}  \left\| \mathbf{c}_i - \mathbf{c}_{i-1} \right\|_2^2 \end{equation} N represents the number of control points in the B-spline interpolation.

    \item \textbf{Centrality Loss}: Penalizes deviations of the control points from a central value (e.g., 0).  \begin{equation}\mathcal{L}_{\text{Centrality}} = \sum_{i=1}^N  \left\| \mathbf{c}_i \right\|_2^2 \end{equation}
    
    \item \textbf{Mean Bias Loss}: Penalizes deviations of the mean layer-wise bias from zero to encourage balanced adjustments across layers.  \begin{equation}\mathcal{L}_{\text{MeanBias}} = \mu _b^2 \end{equation} where $\mu_b^2$ is the mean of the layer-wise bias.

    \item \textbf{Variance Bias Loss}: Encourages the variance of the layer-wise bias to match a desired target variance $\sigma_{target}^2$.  \begin{equation}\mathcal{L}_{\text{VarianceBias}} = (\sigma_b^2 - \sigma_{target}^2)^2 \end{equation} where $\sigma_b^2$ is the variance of the layer-wise bias.
\end{itemize}
The total alpha regularization loss is:

\begin{equation}\mathcal{L}_{AlphaReg} = \lambda_{Smoothness}\mathcal{L}_{Smoothness} +  \lambda_{Centrality}\mathcal{L}_{Centrality} + \lambda_{MeanBias}\mathcal{L}_{MeanBias} +\lambda_{VarianceBias}\mathcal{L}_{VarianceBias} \end{equation}

\paragraph{The Total Losses}
The total 1D loss combines the language modeling loss and the reconstruction loss. Each component is weighted to balance their contributions during training.
\begin{equation}
\mathcal{L_\text{1D}} = \lambda_{\text{Recon}} \mathcal{L}_{\text{Recon}} + \lambda_{\text{LM}} \mathcal{L}_{\text{LM}}
\end{equation}
The total 2D loss combines all the aferomentionned losses weighted to balance their contributions during training.

\begin{equation}
\mathcal{L_\text{2D}} = \lambda_{\text{Recon}} \mathcal{L}_{\text{Recon}} + \lambda_{\text{LM}} \mathcal{L}_{\text{LM}} + \lambda_{\text{AlphaReg}} \mathcal{L}_{\text{AlphaReg}} 
\end{equation}

\subsection{1D-alpha Model Training algorithm}

\begin{algorithm}[H]
\caption{Merging Two LLMs with B-spline Weight Blending and Autoencoders}
\label{alg:merge_llms}
\begin{algorithmic}[1]
\REQUIRE Base model parameters $\{\theta_l^{\text{base}}\}_{l=1}^L$, fine-tuned model parameters $\{\theta_l^{\text{fine}}\}_{l=1}^L$, number of layers $L$, number of control points $N$, B-spline degree $k$, control point parameters $\{c_i\}_{i=1}^N$, autoencoder parameters $\{\phi_l\}_{l=1}^L$
\ENSURE Merged model parameters $\{\theta_l\}_{l=1}^L$, trained autoencoders $\{\text{AE}_l\}_{l=1}^L$
\STATE \textbf{Compute blending coefficients} $\alpha(l)$ for each layer $l$ using B-spline interpolation:
\FOR{$l = 1$ to $L$}
    \STATE $\alpha(l) \gets \text{clamp}\left( \sum_{i=1}^{N} c_i B_{i,k}(l) + b_l,\ 0,\ 1 \right)$
\ENDFOR
\STATE \textbf{Merge model parameters} using the blending coefficients:
\FOR{$l = 1$ to $L$}
    \STATE $\theta_l \gets (1 - \alpha(l)) \theta_l^{\text{base}} + \alpha(l) \theta_l^{\text{fine}}$
\ENDFOR
\STATE \textbf{Initialize} autoencoders $\{\text{AE}_l\}_{l=1}^L$ with parameters $\{\phi_l\}_{l=1}^L$
\STATE \textbf{Training Phase}:
\WHILE{not converged}
    \STATE Sample a minibatch of inputs $\mathbf{x}$ and targets $\mathbf{y}$
    \STATE \textbf{Forward pass} through the merged model:
    \STATE $\mathbf{h}_0 \gets \text{Embed}(\mathbf{x})$
    \FOR{$l = 1$ to $L$}
        \STATE $\mathbf{h}_l \gets \text{TransformerLayer}(\hat{\mathbf{h}}_{l-1}; \theta_l)$
        \STATE \textbf{Autoencoder processing at layer} $l$:
        \STATE \quad \textbf{Encoder}:
        \STATE \quad $\mathbf{z}_l \gets \sigma_{\text{enc}}(f^{1D}_{enc}( \mathbf{h}_l ))$
        \STATE \quad \textbf{Decoders}:
        \STATE \quad $\hat{\mathbf{h}}_l \gets f^{1D}_{dec}( \mathbf{z}_l )$
    \ENDFOR
    \STATE \textbf{Compute output logits}:
    \STATE $\text{logits} \gets \text{LMHead}(\hat{\mathbf{h}}_L)$
    \STATE \textbf{Compute losses}:
    \STATE \quad $\mathcal{L}_{\text{LM}} \gets \text{CrossEntropyLoss}(\text{logits}, \mathbf{y})$
    \STATE \quad $\mathcal{L}_{\text{Recon}} \gets \sum_{l=1}^L\left[ \lambda_{\text{MSE}}\mathbb{E}\left[ \left\| \hat{\mathbf{h}}_l - \mathbf{h}_l^{\text{target}} \right\|_2^2\right] + \lambda_{\text{L2}} \left\| \hat{\mathbf{h}}_l - \mathbf{h}_l^{\text{target}} \right\|_2\right] $

    \STATE \textbf{Compute total loss}:
    \STATE $\mathcal{L_\text{1D}} \gets \lambda_{\text{LM}} \mathcal{L}_{\text{LM}} + \lambda_{\text{Recon}} \mathcal{L}_{\text{Recon}}$
    \STATE \textbf{Backpropagate and update} control points $\{c_i\}$ and autoencoder parameters $\{\phi_l\}$
\ENDWHILE
\STATE \textbf{Inference Phase}:
\STATE Use merged model parameters $\{\theta_l\}$ and autoencoders $\{\text{AE}_l\}$ for inference
\end{algorithmic}
\end{algorithm}
\subsection{2D-alpha Model Training algorithm}

\begin{algorithm}[H]
\caption{Merging Two LLMs with 2D B-spline Weight Blending and Dual-Pathway Autoencoders}
\label{alg:merge_llms_2d}
\begin{algorithmic}[1]
\REQUIRE Base model parameters $\{\theta_{l}^{\text{base}}\}_{l=1}^{L}$, fine-tuned model parameters $\{\theta_{l}^{\text{fine}}\}_{l=1}^{L}$, number of layers $L$, number of control points $N$, B-spline degree $k$, control point parameters $\{c_{i}\}_{i=1}^{N}$, autoencoder parameters $\{\phi_l\}_{l=1}^L$
\ENSURE Merged model parameters $\{\theta_{l}\}_{l=1}^{L}$, trained autoencoders $\{\text{AE}_l\}_{l=1}^L$
\STATE \textbf{Compute dimension-wise blending coefficients} $\alpha(l)$ for each layer $l$ and vector $V_d$:
\FOR{$l = 1$ to $L$}
    
        \STATE $\alpha(l) \gets \text{clamp}\left( \sum_{i=1}^N c_{i}\odot B_{i,k}(l) + b_{l},\ 0,\ 1 \right)$
    
\ENDFOR
\STATE \textbf{Merge model parameters} using the blending coefficients:
\FOR{$l = 1$ to $L$}
         \STATE $\theta_{l} = ( \mathbbm{1} - \alpha(l))\odot \theta_{l}^{\text{base}} + \alpha(l)  \odot\theta_{l}^{\text{fine}}$
\ENDFOR
\STATE \textbf{Initialize} autoencoders $\{\text{AE}_l\}_{l=1}^L$ with parameters $\{\phi_l\}_{l=1}^L$
\STATE \textbf{Training Phase}:
\WHILE{not converged}
    \STATE Sample a minibatch of inputs $\mathbf{x}$ and targets $\mathbf{y}$
    \STATE \textbf{Forward pass} through the merged model:
    \STATE $\mathbf{h}_0 \gets \text{Embed}(\mathbf{x})$
    \FOR{$l = 1$ to $L$}
        \STATE $\mathbf{h}_l \gets \text{TransformerLayer}(\hat{\mathbf{h}}_{l-1}; \theta_l)$
        \STATE \textbf{Autoencoder processing at layer} $l$:
        \STATE \quad \textbf{Local Pathway (Convolutions)}:
        \STATE \quad $\mathbf{z}_l^{\text{local}} \gets \sigma_{\text{local}}\left(f^{2D}_{local}(\mathbf{h}_l)\right)$
        \STATE \quad \textbf{Global Pathway (Residual Adapter)}:
        \STATE \quad $\mathbf{z}_l^{\text{global}} \gets f^{2D}_{global}(\mathbf{h}_l)$
        \STATE \quad \textbf{Combine Bottleneck Representations}:
        \STATE \quad $\mathbf{z}_l \gets \text{Concat}(\mathbf{z}_l^{\text{local}}, \mathbf{z}_l^{\text{global}})$
        \STATE \quad \textbf{Decoder Reconstruction}:
        \STATE \quad $\hat{\mathbf{h}}_l \gets f^{2D}_{dec}(\mathbf{z}_l)$
    \ENDFOR
    \STATE \textbf{Compute output logits}:
    \STATE $\text{logits} \gets \text{LMHead}(\hat{\mathbf{h}}_L)$
    \STATE \textbf{Compute losses}:
    \STATE \quad $\mathcal{L}_{\text{LM}} \gets \text{CrossEntropyLoss}(\text{logits}, \mathbf{y})$
    \STATE \quad $\mathcal{L}_{\text{Recon}} \gets \sum_{l=1}^L\left[ \lambda_{\text{MSE}}\mathbb{E}\left[ \left\| \hat{\mathbf{h}}_l - \mathbf{h}_l^{\text{target}} \right\|_2^2\right] + \lambda_{\text{L2}} \left\| \hat{\mathbf{h}}_l - \mathbf{h}_l^{\text{target}} \right\|_2\right]  $
    \STATE \quad $\mathcal{L}_{\text{AlphaReg}} \gets \lambda_{\text{Smoothness}} \mathcal{L}_{\text{Smoothness}} + \lambda_{\text{Centrality}} \mathcal{L}_{\text{Centrality}} + \lambda_{\text{MeanBias}} \mathcal{L}_{\text{MeanBias}} + \lambda_{\text{VarianceBias}} \mathcal{L}_{\text{VarianceBias}}$
    \STATE \textbf{Compute total loss}:
    \STATE $\mathcal{L_\text{2D}} \gets \lambda_{\text{LM}} \mathcal{L}_{\text{LM}} + \lambda_{\text{Recon}} \mathcal{L}_{\text{Recon}} + \lambda_{\text{AlphaReg}} \mathcal{L}_{\text{AlphaReg}}$
    \STATE \textbf{Backpropagate and update} control points $\{c_{i}\}$ and autoencoder parameters $\{\phi_l\}$
\ENDWHILE
\STATE \textbf{Inference Phase}:
\STATE Use merged model parameters $\{\theta_{l}\}$ and autoencoders $\{\text{AE}_l\}$ for inference
\end{algorithmic}
\end{algorithm}

\section{Additional results of the 1D-alpha Model experiment}

\subsection{Visualization of Hidden States in Layer 4}

To further illustrate how the autoencoders enable the superposition of representations, we perform a t-SNE visualization of the hidden states from layer 4. Layer 4 is chosen as an early intermediate feature specialized block.

\begin{figure}[H]
\centering
\includegraphics[width=0.6\textwidth]{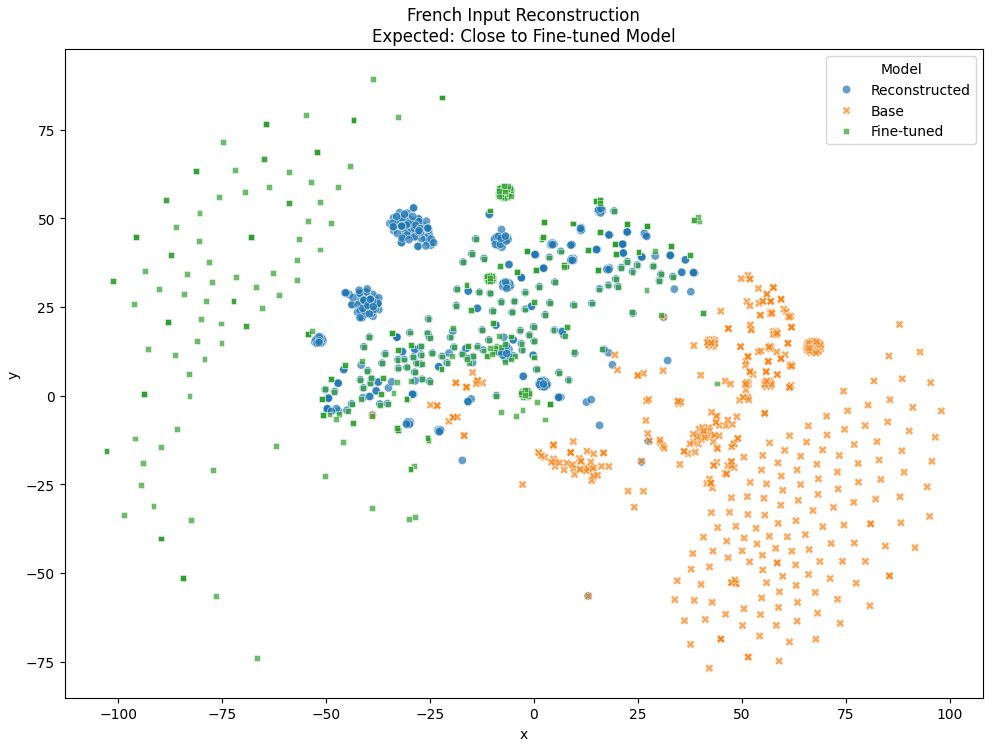}
\caption{t-SNE visualization of layer 4 hidden states from the merged model and expert models for French inputs. }
\end{figure}

\begin{figure}[H]
\centering
\includegraphics[width=0.6\textwidth]{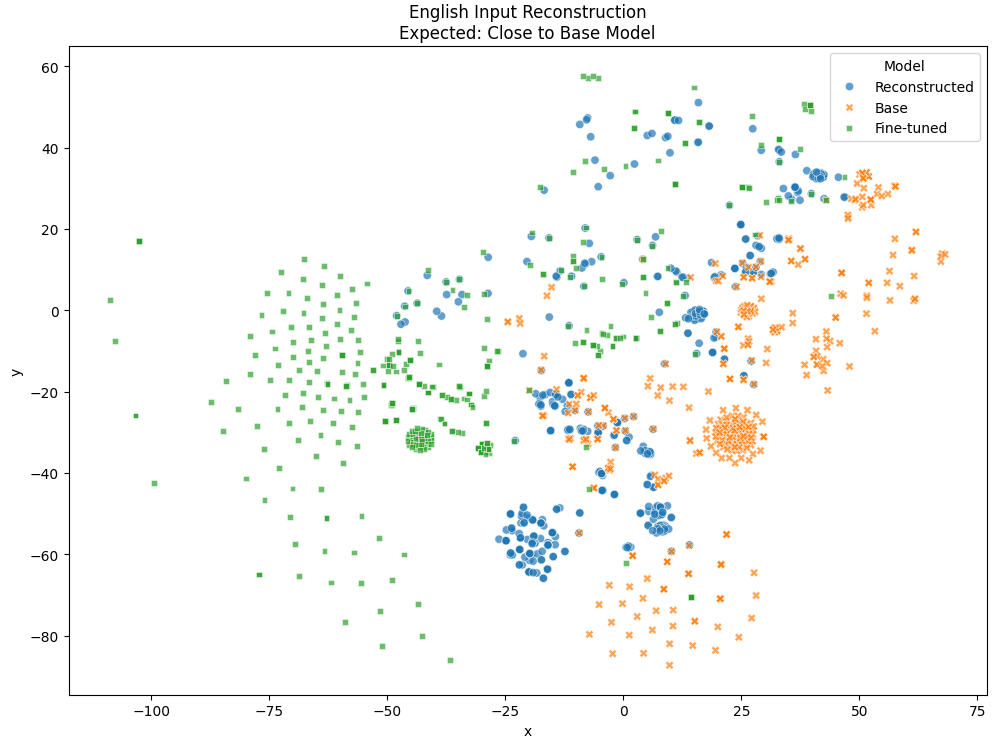}
\caption{t-SNE visualization of layer 4 hidden states from the merged model and expert models for English inputs.}
\end{figure}

\paragraph{Analysis}

The t-SNE projection in Figures 6 and 7 provide a visual representation of how the merged model's layer 4 hidden states are reconstructed based on the input data. Specifically:

\begin{itemize}
    \item \textbf{English Inputs}: The reconstructed hidden states of the merged model (blue points) cluster closely with the hidden states of the base model (green points). This indicates that for English inputs, the autoencoders adjust the blended hidden states to align with the base model's representations.
    \item \textbf{French Inputs}: The reconstructed hidden states of the merged model (red points) cluster near the hidden states of the fine-tuned model (orange points). This shows that for French inputs, the autoencoders reconstruct the hidden states to resemble those of the fine-tuned model.
\end{itemize}

This behavior demonstrates that the autoencoders effectively condition the merged model's internal representations based on the input language. By reconstructing the hidden states to match the appropriate expert model, the merged model can adaptively process different languages within a unified framework.

\paragraph{Implications for Superposition}

The t-SNE visualization highlights the role of autoencoders in enabling superposition:

\begin{itemize}
    \item \textbf{Adaptive Representation}: The merged model produces hidden states that are contextually aligned with the relevant expert model, showcasing its ability to adapt representations based on input data.
    \item \textbf{Efficient Parameter Usage}: By superimposing the representations within the same parameters, the model avoids redundancy and efficiently utilizes its capacity.

\end{itemize}

\subsection{Evaluation of Hidden State Trajectories}
To evaluate the behavior of the 1D Merged Model, we analyzed the trajectories of hidden states across selected layers in both English and French language samples. This analysis helps to visually understand how the merged model acts given different input.
Principal Component Analysis (PCA) was used to project the high-dimensional hidden states into a 2D space, preserving the primary variance for visualization. The hidden states were extracted for layers 3, 4, 5, 6, and 7 from the base, fine-tuned, and merged models so the markers in figure (8) indicated the start (layer 3) and end (layer 7) points, and annotations denoted the layers.

\begin{figure}[H]  
\centering
\includegraphics[width=1\textwidth]{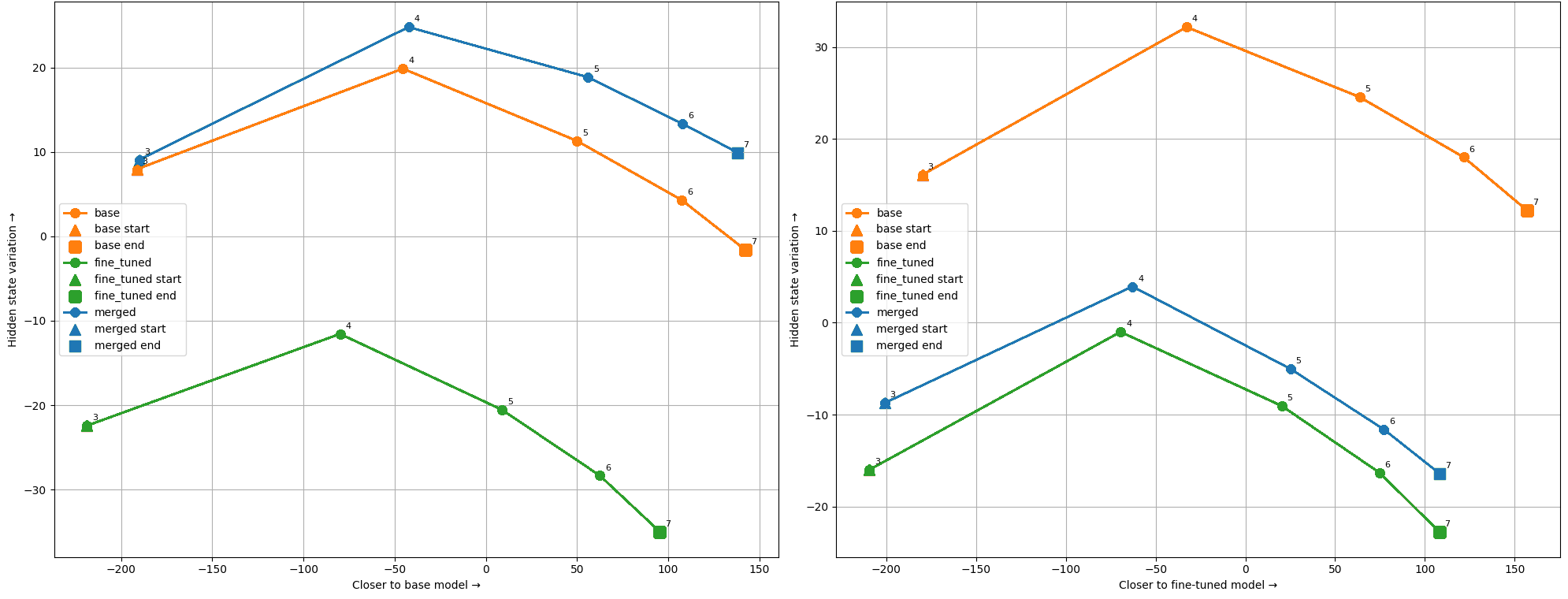}
\caption{PCA visualization of hidden state trajectories from the base, fine-tuned, and merged models across layers 3 to 7 for English and French inputs.}
\end{figure}


\end{document}